\documentclass[letterpaper, 10 pt, conference]{ieeeconf} 

\IEEEoverridecommandlockouts                              

\usepackage[T1]{fontenc}

\usepackage{amsmath} 
\usepackage{amssymb}  
\usepackage{graphicx} 
\usepackage{ragged2e}
\usepackage{booktabs}
\let\labelindent\relax
\usepackage{enumitem}

\usepackage{siunitx}
\usepackage[ruled,vlined]{algorithm2e}
\usepackage{algorithmic}
\usepackage{microtype}
\usepackage{flushend}
\usepackage[hang,flushmargin]{footmisc}

\usepackage{url}
\usepackage[table, dvipsnames]{xcolor}
\makeatletter
\g@addto@macro{\endtabular}{\rowfont{}}
\makeatother
\newcommand{\rowfonttype}{}
\newcommand{\rowfont}[1]{
\gdef\rowfonttype{#1}#1\ignorespaces%
}
\makeatother
\usepackage[export]{adjustbox}

\makeatletter
\let\NAT@parse\undefined
\makeatother

\usepackage{pifont}

\usepackage{siunitx}
\usepackage{cuted}
\usepackage{threeparttable}
\usepackage{multirow}
\usepackage{cite}

\usepackage{tikz}
\usepackage{pgfplots}
\usepackage[pdfencoding=auto, colorlinks=true]{hyperref} 

\usepackage[capitalize]{cleveref}
\crefname{section}{Sec.}{Secs.}
\Crefname{section}{Section}{Sections}
\Crefname{table}{Table}{Tables}
\crefname{table}{Tab.}{Tabs.}

\usepackage[resetlabels]{multibib}
\newcites{S}{References}

\usepackage{tabularx}
\newcolumntype{Y}{>{\centering\arraybackslash}X}
\newcolumntype{Z}{>{\raggedleft\arraybackslash}X}

\definecolor{dark-green}{RGB}{12,80,12}

\renewcommand{\eqref}[1]{Eq.~(\ref{#1})}

\newcommand{\ours}{LoopGNN}

\renewcommand{\baselinestretch}{0.985}

\newcommand{\greyrule}{\arrayrulecolor{black!30}\midrule\arrayrulecolor{black}}

\title{\LARGE \bf
Visual Loop Closure Detection Through Deep Graph Consensus
}

\author{Martin Büchner$^{1}$, Liza Dahiya$^{2}$, Simon Dorer$^{1}$, Vipul Ramtekkar$^{2}$,  \\[4pt] Kenji Nishimiya$^{2}$, Daniele Cattaneo$^{1}$, Abhinav Valada$^{1}$
\thanks{$^{1}$ Department of Computer Science, University of Freiburg, Germany.}%
\thanks{$^{2}$ Solution System Development Center in Honda R\&D Co,. Ltd., Japan.}%
\thanks{This paper provides a supplementary material at \url{https://loopgnn.cs.uni-freiburg.de}. This work was supported by Honda R\&D Co., Ltd., an academic grant from NVIDIA and the BrainLinks-BrainTools Center of the University of Freiburg. We sincerely thank Fabian Schmidt and Nick Heppert for discussions.}
}

\begin{document}

\maketitle
\thispagestyle{empty}
\pagestyle{empty}

\begin{abstract}
Visual loop closure detection traditionally relies on place recognition methods to retrieve candidate loops that are validated using computationally expensive RANSAC-based geometric verification.  As false positive loop closures significantly degrade downstream pose graph estimates, verifying a large number of candidates in online simultaneous localization and mapping scenarios is constrained by limited time and compute resources.
While most deep loop closure detection approaches only operate on pairs of keyframes, we relax this constraint by considering neighborhoods of multiple keyframes when detecting loops. In this work, we introduce LoopGNN, a graph neural network architecture that estimates loop closure consensus by leveraging cliques of visually similar keyframes retrieved through place recognition. By propagating deep feature encodings among nodes of the clique, our method yields high precision estimates while maintaining high recall. Extensive experimental evaluations on the TartanDrive 2.0 and NCLT datasets demonstrate that LoopGNN outperforms traditional baselines. Additionally, an ablation study across various keypoint extractors demonstrates that our method is robust, regardless of the type of deep feature encodings used, and exhibits higher computational efficiency compared to classical geometric verification baselines. We release our code, supplementary material, and keyframe data at \mbox{\url{https://loopgnn.cs.uni-freiburg.de}}.
\end{abstract}

\section{Introduction}%
Simultaneous localization and mapping (SLAM) is a fundamental task in mobile robotics, serving as the foundation for various downstream components such as navigation~\cite{werby24rss,honerkamp2023n} and task planning~\cite{honerkamp2024language}. While odometry methods provide local estimates of robot pose changes~\cite{wang2021tartanvo,vodisch2023covio}, they are inherently subject to accumulating errors, leading to considerable drift and causing downstream tasks to fail due to map inconsistencies. Loop closure detection aims to address this problem by identifying previously visited places and estimating the relative transformation between the current image and the past traversal. This transformation serves as a \textit{hard} constraint throughout pose graph optimization, thereby reducing drift and mitigating errors stemming from noisy odometry. 

\begin{figure}[h]
    \centering
    \includegraphics[width=\linewidth]{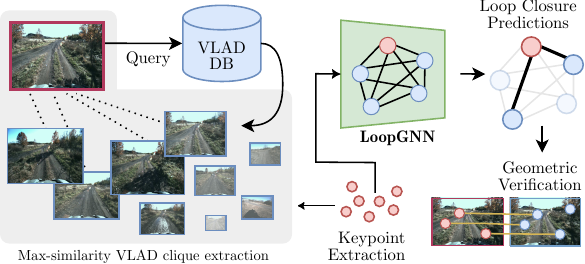}
    \caption{Given a query keyframe, the set of closest keyframes is retrieved through classical place recognition methods such as VLAD~\cite{arandjelovic2013all}. In this work, we introduce LoopGNN, a novel graph neural network that produces precise loop closure estimates via deep graph consensus among the obtained closest set of keyframes.  We find that our method scales across various types of deep keypoint encodings and outperforms baselines relying on only pairs of keyframes to identify loop closures.}
    \label{fig:teaser}
    \vspace*{-.5cm}
\end{figure}

In practice, the majority of established loop closure detection approaches integrated into visual SLAM rely on image retrieval methods~\cite{schmidt2024visual} that are either vocabulary-based, such as DBoW~\cite{galvez2012bags}, or global descriptor-based, such as the family of VLAD feature approaches~\cite{arandjelovic2016netvlad, arandjelovic2013all}. Broadly, these approaches fall under the umbrella of visual place recognition (VPR). Given a query keyframe, those methods provide sets of similar camera images that may represent potential loop closures. As such, VPR approaches are evaluated in terms of high recalls. In order to construct actual loop closure constraints, those proposals require filtering to reduce false positives via compute-intense geometric verification. The most common approach for this task is the RANdom SAmple Consensus (RANSAC)~\cite{fischler81ransac} algorithm under an epipolar constraint~\cite{hartley2003multiple}. Due to its iterative formulation, RANSAC becomes computationally expensive when evaluating a significant number of loop closure candidates, effectively limiting its deployment in real-time applications~\cite{yu2024gv}. 
Additionally, we observe that while deep learning has been successfully employed for place recognition, this is not the case for robust loop closure detection~\cite{tsintotas2022revisiting}. 
This is primarily due to visual loop closure detection being heavily affected by perceptual aliasing and being rather sensitive to varying camera poses. However, we argue that previous deep loop closure detection methods would benefit from not considering only pairs of frames~\cite{zhang2017loop, merrill2018lightweight, yu2024gv, buchner20223d}, but larger neighborhoods.\looseness=-1

In this work, we present a novel deep loop closure detection approach that operates on sets of extracted keyframes given a query frame and estimates loop closures through deep graph consensus. By identifying maximum-similarity cliques given a query frame and its most-similar keyframes, we construct graphs connecting all retrieved frames among each other. We use per-frame keypoint features as node features in our graph and propose a novel graph neural network (GNN) architecture, termed \ours{}, that performs neural message passing to allow information exchange among retrieved keyframes. By combining retrieved keyframes, we aim to learn a deep graph consensus under the assumption that similar locations are retrieved as groups of frames during place recognition. We demonstrate that this ultimately increases the precision of predicted loop closures and decreases the number of pairs that need to undergo geometric verification. 

We demonstrate the effectiveness of our approach on a specifically curated split of the outdoor all-terrain vehicle driving dataset {TartanDrive 2.0}~\cite{tartandrive2} featuring largely unstructured environments that induce a significant degree of perceptual aliasing. Second, we evaluate on the NCLT dataset~\cite{carlevaris2016nclt} and ablate the main design choices of our proposed framework. As part of this work, we make the following contributions:
\begin{itemize}
    \item We propose graph similarity-based consensus estimates for precise loop closure identification through graph neural networks and couple this with geometric verification.
    \item We provide extensive comparisons against previous place recognition and learning-based loop closure detection approaches and delineate the computational benefits of our method over previous work.
    \item We establish a split of the \mbox{TartanDrive\,2.0} dataset that paves the way forward for research on visual-inertial SLAM under dynamic motion in unstructured environments.
    \item We make the reprocessed datasets, the learning framework, and code publicly available at \mbox{\url{https://loopgnn.cs.uni-freiburg.de}}.
\end{itemize}

\section{Related Work}
\label{sec:related-work}
In this section, we review related work on visual correspondence, visual loop closure detection, and graph-based methods for loop closure detection.

{\parskip=3pt
\noindent\textit{Visual Correspondence}: 
Traditional methods for visual correspondence rely on handcrafted feature descriptors such as ORB~\cite{rublee2011orb}, SIFT~\cite{lowe2004sift}, and SURF~\cite{bay2006surf}. These approaches often struggle in environments with occlusions, illumination changes, and strong viewpoint variations. To address these challenges, deep learning-based feature descriptors have emerged as a powerful alternative. LIFT~\cite{yi2016lift} was among the first frameworks to jointly optimize keypoint detection and descriptor extraction. Building on earlier deep learning approaches, SuperPoint~\cite{detone2018superpoint} introduced a fully convolutional network for self-supervised feature extraction that is used by SuperGlue~\cite{sarlin2020superglue} to refine feature matching using graph neural networks. 
Recent methods such as XFeat~\cite{potje2024xfeat} and RoMa~\cite{edstedt2024roma} improve efficiency and robustness in challenging conditions. XFeat uses a lightweight CNN for fast keypoint extraction, while RoMa leverages DiNOv2~\cite{oquab2023dinov2} for accurate dense feature matching in complex environments.}

{\parskip=3pt
\noindent\textit{Visual Loop Closure Detection}: 
Visual loop closure detection often relies on image retrieval techniques where a single query image is matched against a database of previously observed locations. Traditional methods for visual loop closure rely on BoW (Bag-of-Words) or DBoW2~\cite{galvez2012bags}, which approximate nearest neighbors in the image feature space and construct a visual vocabulary for place recognition. Another family of approaches utilizes global descriptor-based methods, where an entire image is represented by a compact vector for efficient matching. A well-known method in this family is VLAD~\cite{arandjelovic2013all}, which aggregates local feature descriptors into a fixed-length representation and tends to be more robust than BoW methods. NetVLAD~\cite{arandjelovic2016netvlad} extends VLAD by incorporating a learnable feature aggregation mechanism using deep networks, improving performance in challenging situations. However, the aforementioned approaches usually require computationally heavy geometric verification through RANSAC~\cite{fischler81ransac}.
}

Place recognition techniques are optimized for high recall values and thus retrieve a broad set of candidate matches to maximize coverage~\cite{arandjelovic2016netvlad, Berton_CVPR_2022_CosPlace, keetha2023anyloc}. However, for robust localization, high precision is crucial to avoid incorrect loop closures that can introduce errors into the pose graph. Only few learning-based methods attempt to directly address the problem of loop closure detection by not depending on prior place recognition methods.
For instance, Calc~\cite{merrill2018lightweight} employs a self-supervised autoencoding scheme where random projective transformations are applied to camera images, and a neural network is trained to predict Histogram of Oriented Gradient (HOG) descriptors of the original image given the augmented one.
Calc2.0~\cite{merrill2019calc2} additionally includes a variational autoencoder and a triplet-loss embedding network to embed images and cluster same-place images together. Siamese-ResNet~\cite{qiu2018siamese}, on the other hand, employs a Siamese network to train a contrastive loss for robust image similarity learning. Finally, LoopCNN~\cite{zhang2017loop} extracts high-dimensional image descriptors using a pre-trained CNN and compares PCA-reduced features across image pairs.

{\parskip=3pt
\noindent\textit{Graph Estimation for Loop Closure Detection}: 
Various methods have investigated graph structures to describe either the scene morphology or objects including their semantics depicted in images~\cite{yu2022semanticloop, an2019fast, qian2022towards, duan2022deep, kim2022closing} in order to find feature-rich representations for matching.
Another branch of work entirely focuses on modeling ambiguous observations directly rather than relying on post-hoc corrections by employing graph-based probabilistic reasoning. This enables the identification of perceptual aliases, thereby reducing false positive loop closures~\cite{suenderhauf2012switchable, lajoie2019modeling}.
In contrast to the aforementioned works, our approach focuses on employing deep graph structures at the intersection of visual place recognition and loop closure detection. In detail, we consider graphs of retrieved images given a query image in order to learn consensus estimates for loop closure detection.
\section{LoopGNN Method}
In this section, we detail our proposed LoopGNN approach that performs loop closure detection by estimating deep graph consensus and predicting relative transformations. Our approach consists of a method for extracting similarity cliques given a query keyframe and a graph neural network for encoding keyframe features as detailed in \cref{fig:overview}.

\begin{figure*}
    \centering
    \includegraphics[width=1.0\linewidth, trim=0.0cm 0.0cm 0.0cm 0.0cm, clip]{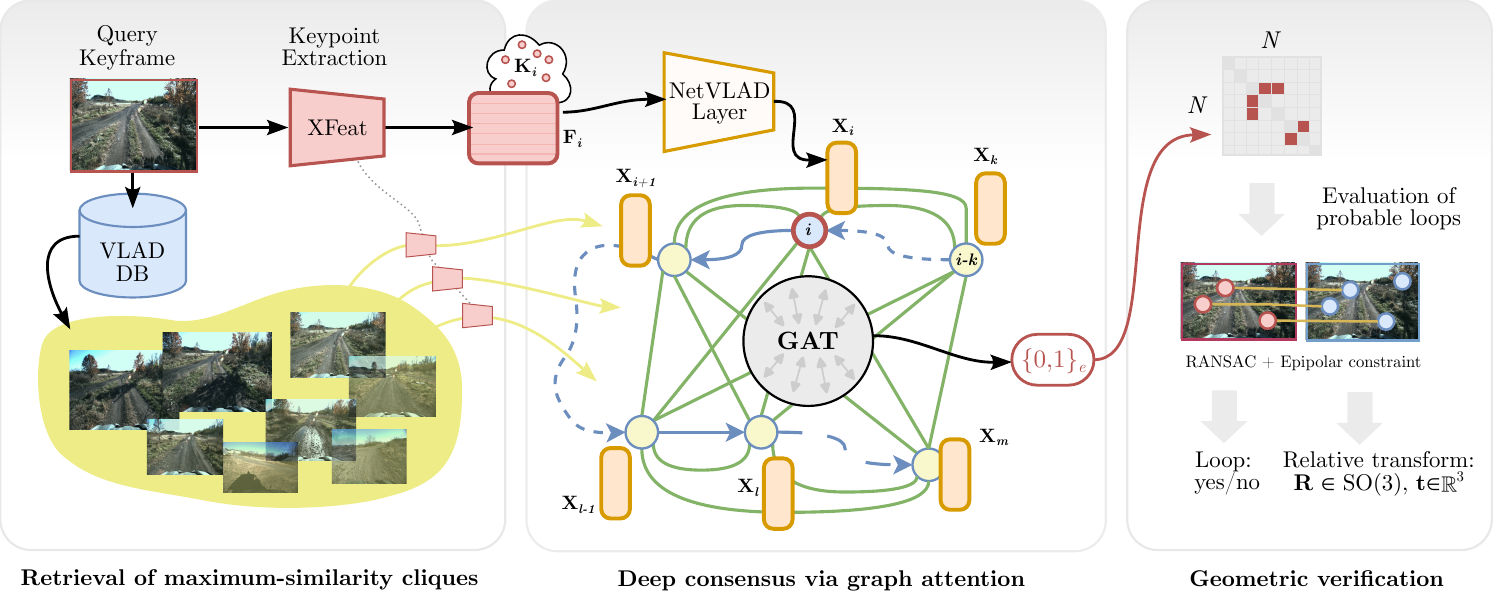}
    \caption{Overview of our LoopGNN approach: We create keyframes from robot trajectories and utilize a deep keypoint extractor such as XFeat~\cite{potje2024xfeat} to obtain keypoints for each image. Next, we fit a VLAD-based place recognition model allowing robust and fast retrieval of similar frames given a query frame (left). In the following, given a query frame, we independently encode the keypoint descriptors of all frames (query and retrieved ones) using a NetVLAD layer and construct a neighborhood graph. We feed this attributed graph into a graph attention network in order to produce a deep consensus regarding loop closures among keyframes of the neighborhood (middle). Finally, we extract the set of highest-scoring edge-wise predictions of the network and validate pairs of frames using RANSAC-based geometric verification (right).
    } 
    \label{fig:overview}
    \vspace*{-.3cm}
\end{figure*}

\subsection{Keyframe Extraction}
\label{sec:feat-keyframe-ext}
In order to sample keyframes, we follow previous work~\cite{qin2019b} that extracts frames when a cumulative distance threshold of \SI{0.5}{m} is exceeded. For each sequence, we obtain a posed graph of keyframes  $\mathcal{G}_K=\{V_i \}, $ where $i \in \{1,2,\ldots,N\}$, and $N$ is the total number of keyframes in the sequence.
For each of those keyframes $V_i$ we assume access to its camera image $\mathbf{I}_{i} \in \mathbb{R}^{H \times W \times C}$. Next, we feed $\mathbf{I}_{i}$ through any kind of lightweight keypoint feature extractor $f_{\text{KP}}(\,\mathbf{I}_i\,) = \{\mathbf{K}_i, \mathbf{F}_i\}$, which provides us with a set of keypoints $\mathbf{K}_i \in \mathbb{R}^{N_{K} \times 2}$ and their associated keypoint descriptors $\mathbf{F}_{i} = \left[ \mathbf{f}_{i,1}, \mathbf{f}_{i,2}, \ldots, \mathbf{f}_{i,N_K}\right] \in \mathbb{R}^{{N_K} \times {N_{KD}}}$, where $N_K$ may vary per image.

\subsection{Global Feature Aggregation}
\label{sec:method-global-feat-agg}
While the obtained set of keypoint descriptors represents a reliable basis for re-identification, it is generally too high-dimensional to be used for online similarity search. In order to facilitate fast similarity search over $\mathcal{G}_{K}$, we employ the Vector-of-Linearly-Aggregated-Descriptors (VLAD) technique. By clustering the feature descriptors $\mathbf{F}_{i}$ of the generated keypoints using $N_C$ cluster prototypes we construct a vocabulary that allows the generation of image-level features. Since the keypoints may stem from non-Euclidean manifolds, we employ either a cosine similarity-based or Euclidean distance-based variant. We obtain image-level VLAD features by computing the sum of residuals of the keypoint descriptors of an image to each of the cluster centers $\mathbf{c}_j$, with $j \in \{1,2,\ldots,N_C\}$, as:
\begin{equation}
    \mathbf{D}_{i,,j} = \sum_{k=1}^{N_{K}} \beta_j (\mathbf{f}_{i,k}) \cdot \operatorname{cos}(\mathbf{f}_{i,k}, \mathbf{c}_j),
\end{equation}
where in the case of hard-assignment, that we employ, $\beta_j$ equals 1 if $\mathbf{f}_{i,k}$ is assigned to $\mathbf{c}_j$ and 0 otherwise. Finally, we craft image-level VLAD descriptors $\mathbf{D}_{i}$ via concatenation of cluster-aggregated descriptors and normalization as detailed in prior work~\cite{arandjelovic2013all}. Thus, each of the keyframes $V_i \in \mathcal{G}_K$ is parametrized as follows:
\begin{equation}
    V_i = \{\mathbf{I}_i,\, \mathbf{F}_{i} ,\, \mathbf{K}_i,\, \mathbf{D}_{i}\}.
\end{equation}

\subsection{Graph Construction}
In the following, we define \textit{maximum-similarity cliques} by performing a dense similarity search across all global keyframes descriptors $\mathbf{D}_{i}$. This provides us with a set of maximum-similar keyframes given a query keyframe $q$. At inference time, we query the existing keyframes' VLAD descriptors based on the query descriptor using cosine similarities. Next, we identify the closest $k$-\% of keyframes to the query keyframe and construct an interconnected, undirected graph of keyframes $\mathcal{G}(i=q) = (V_q, E_q)$ that constitutes a maximum-similarity clique. Given the query keyframe, we partition all created edges given their incidence to the query keyframe and define \textit{query edges} as depicted in Fig.~\ref{fig:neighborhood-construction}. In the following, we detail how to obtain a deep consensus estimate for improved loop closure detection by passing messages within the clique.

\subsection{Graph Attention Network}
\label{sec:message-passing}
As part of this work, we propose a novel approach that provides a graph consensus by learning expressive node features from keypoint descriptors. To do so, we require an order-invariant operation that aggregates descriptors. Given the successful use of the VLAD clustering mentioned above, we employ its learned counterpart, namely NetVLAD embeddings~\cite{arandjelovic2016netvlad}, to aggregate keypoint descriptors $\mathbf{F}_{i}$ into preliminary node features, see Fig.~\ref{fig:overview}. In the next step, we downproject the NetVLAD features to form node features $\mathbf{X}= [\mathbf{X}_1, \ldots, \mathbf{X}_N]^T \in \mathbb{R}^{N\times 256} $ and perform a series of message passing steps. Throughout our experiments, we found that the graph attention mechanism (GAT)~\cite{velickovic2018graph} performs best in learning descriptive node features through graph interactions, ultimately providing more reliable loop closure predictions. Within each GAT layer, learned attention scores are estimated to produce weighted transformations of the original node features as follows:
\begin{equation}
    \mathbf{X}_m^{(l+1)} = \sigma\left(\sum_{n \in \mathcal{N}(m)} \alpha_{mn}\mathbf{W}^{(l)}\mathbf{X}_n^{(l)}\right),
\end{equation}
where $\mathbf{X}_m^{(l+1)}$ is the updated node feature, $\mathbf{W}^{(l)}$ is the learnable weight matrix of the involved layer $l$, $\mathcal{N}(m)$ represents the 1-hop neighborhood of node $m$, and $\alpha_{mn}$ is the attention weight between node $m$ and $n$. We chain multiple GAT layers to iteratively refine node features. After propagating messages for $L$ iterations, we concatenate the node features $\mathbf{X}^{(L)}$ of nodes incident to each edge in order to form edge features that allow regressing binary scores that constitute loop closure predictions. We supervise the network by employing a binary cross-entropy loss, where the ground truth is one if the relative pose between the query image and the considered loop candidate is lower than a threshold of \SI{4}{m} and \SI{30}{\degree}.
We additionally incorporate loop closure candidates across sequences to allow for improved robustness. Furthermore, we found that the accuracy of predictions is particularly high on the query edges as visualized in Fig.~\ref{fig:neighborhood-construction}. Given that finding, we only consider the scores of those edges during inference. The remaining set of edge scores is disregarded and thus only serves the consensus estimate of the query edges.

\begin{figure}
    \centering
    \includegraphics[width=0.9\linewidth]{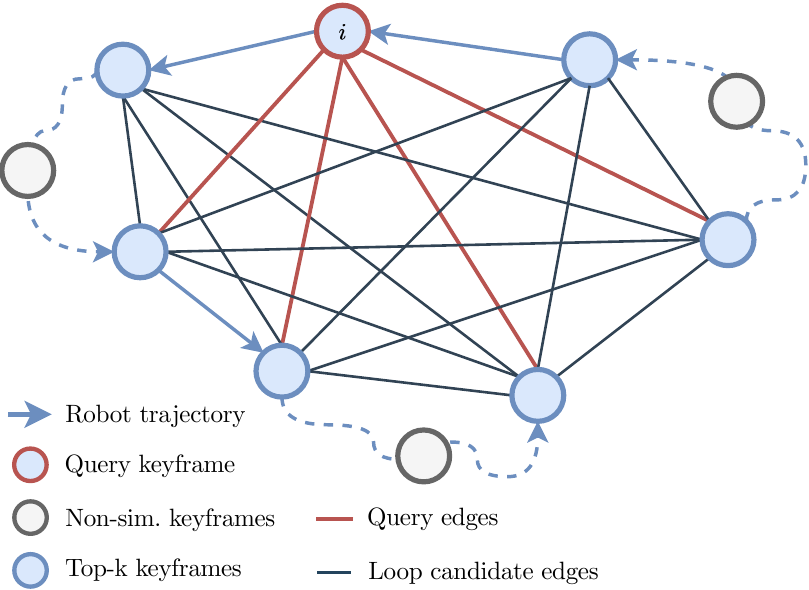}
    \caption{Maximum-similarity clique construction: Given robot trajectory, we extract keyframes at fixed distance intervals. Given a query keyframe, we retrieve the set of top-k closest keyframes using VLAD-based place recognition. We impose interconnected edges and take the obtained graph $\mathcal{G}(q)=(V_q, E_q)$ as input to our LoopGNN pipeline. At the output, we only consider the edge scores of the query edges denoted here.}
    \label{fig:neighborhood-construction}
    \vspace{-0.3cm}
\end{figure}

\subsection{Geometric Verification}
In the following, we employ geometric verification to first, increase robustness of the predicted loop scores, and second, obtain relative camera transforms.
We analyze the score distribution of the GNN outputs and choose a threshold that preserves high recalls across multiple sequences. Next, we validate keyframe pairs $(i,j)$ whose score is higher than the chosen threshold, and retrieve the associated keypoints $\mathbf{K}_{i}, \mathbf{K}_{j}$, along with their corresponding descriptors $\mathbf{F}_{i}, \mathbf{F}_{j}$. Next, we compute mutual keypoint correspondences among the two images. In order to verify whether each candidate loop is geometrically valid, we compute the underlying fundamental matrix $\mathcal{F}$ that satisfies the epipolar constraint:
\begin{equation}
    \hat{\mathbf{K}}_j^{T}\mathcal{F}\hat{\mathbf{K}}_i = 0,
\end{equation}
where $\hat{\mathbf{K}}_i$ and $\hat{\mathbf{K}}_j$ denotes the set of matched keypoints in image $\mathbf{I}_i$ and $\mathbf{I}_j$, respectively. We repeat this optimization iteratively using the robust RANSAC estimator~\cite{fischler81ransac} to obtain the transformation with the highest percentage of inlier keypoints. Given that a particular projection threshold of 50\% is exceeded, we further obtain the associated relative camera transformation, i.e., the relative rotation $\mathbf{R} \in SO(3)$ and translation $\mathbf{t} \in \mathbb{R}^3$ by computing the corresponding essential matrix:
\begin{equation}
    \mathcal{E} = \mathcal{K}_{C}^T \mathcal{F} \mathcal{K}_{C}, 
\end{equation}
where $\mathcal{K}_{C} \in \mathbb{R}^{3\times3}$ denotes the camera intrinsic matrix. Finally, we perform Singular Value Decomposition (SVD) of $\mathcal{E}$ and obtain the corresponding rotational and translational component ($\mathbf{R}, \mathbf{t}$) up to scale.

\section{Experimental Evaluation}\label{sec:experiments}
In this section, we demonstrate the capabilities of our proposed LoopGNN framework on two distinct datasets, \textit{TartanDrive 2.0}~\cite{tartandrive2} and \textit{NCLT}~\cite{carlevaris2016nclt}, and compare it against previous VPR, deep loop closure detection, and classical feature matching and geometric verification methods.

\subsection{Datasets}
\label{sec:exp-datasets}
In the following, we provide details on the datasets that we employ for evaluation:

\noindent{\textit{TartanDrive 2.0}}: The TartanDrive 2.0 dataset~\cite{tartandrive2} (TD2) is a large-scale off-road terrain driving dataset tailored for self-supervised learning of dynamics models. It consists of unstructured and texture-less sequences. Different from its original intended use case, in this work, we propose a dataset split that facilitates research in the domain of visual-inertial SLAM. Specifically, we select sequences that contain sufficient IMU initialization phases at their beginnings where the vehicle does not move and that contain enough loop closures. At the same time, we believe that the amount of perceptual aliasing and high speed driving behavior represents a highly challenging test bed for visual odometry and mapping research.
In this work, we have constructed a train/test-split that consists of 7 training and 3 test scenes, respectively containing 16625 and 23433 keyframes. As described in \cref{sec:feat-keyframe-ext}, we sample keyframes following the procedure introduced in VINS-Fusion~\cite{qin2019b}. We list all considered sequences in \cref{tab:td2-dataset}.

\noindent{\textit{NCLT}}: The NCLT dataset~\cite{carlevaris2016nclt} is a large-scale collection of 27 LiDAR-camera sequences recorded with a mobile robot across indoor and outdoor environments. It contains omnidirectional imagery, 3D LiDAR, IMU, and GNSS measurements, and provides global ground-truth robot trajectories optimized across all sequences. Each sequence covers the same area traversed at different times across all four seasons and features dynamic obstacles, viewpoint variations, and changing lighting conditions. In our experiments, we follow the train/test split of LiSA~\cite{Yang_2024_CVPR} with four sequences used for training and four sequences used for validation. Since the contained sequences are very long, we limit the number of keyframes via position-based cropping to a subset of the scene, which yields between 4000 to 6000 keyframes per sequence under the same keyframe generation scheme as used for TD2. All NLCT sequences used are listed in \cref{tab:nclt-dataset}.

\begin{table}
    \footnotesize
    \setlength{\tabcolsep}{8pt}
    \centering
    \caption{Overview of the selected TartanDrive 2.0 dataset sequences}
    \label{tab:td2-dataset}
    \begin{threeparttable}
    \begin{tabularx}{\linewidth}{lll}
      \toprule
        Split & Sequence & \# Frames \\
        \midrule
        \multirow{7}{*}{Train} & \textit{2023-11-14-14-26-22\_gupta} & 1272 \\  
        & \textit{2023-11-14-14-34-53\_gupta} & 1632 \\
        & \textit{2023-11-14-14-45-20\_gupta} & 823 \\
        & \textit{2023-11-14-14-52-22\_gupta} & 2066 \\
        & \textit{2023-11-14-15-02-21\_figure\_8} & 3782 \\
        & \textit{gupta\_skydio\_2023-09-14-11-03-09} & 853 \\
        & \textit{turnpike\_2023-09-12-12-53-32} & 6197 \\
        \midrule
        \multirow{3}{*}{Test} & \textit{figure\_8\_2023-09-13-17-24-26} & 4429 \\
        & \textit{figure\_8\_morning\_2023-09-12-10-37-17} & 11065 \\
        & \textit{figure\_8\_morning\_slow\_2023-09-12-11-06-32} & 7939 \\
      \bottomrule
    \end{tabularx}
    \begin{tablenotes}[para,flushleft]
       \footnotesize   
       All selected TD2 sequences contain at least one loop and are captured under varying weather conditions. All train sequences cover distinct regions while the test sequences show partial overlap.
     \end{tablenotes}
   \end{threeparttable}
\end{table}

\begin{table}
    \footnotesize
    \setlength{\tabcolsep}{9pt}
    \centering
    \caption{Overview of the NCLT dataset sequences}
    \label{tab:nclt-dataset}
    \begin{threeparttable}
    \begin{tabularx}{\linewidth}{lllll}
      \toprule
        Split & Sequence & \# Frames & Daytime & Weather\\
        \midrule
        \multirow{4}{*}{Train} & \textit{2012-01-22} & 5321 & afternoon & cloudy \\  
        & \textit{2012-02-02} & 5932 & afternoon & sunny \\
        & \textit{2012-02-18} & 5383 & evening & sunny \\
        & \textit{2012-05-11} & 5635 & midday & sunny \\
        \midrule
        \multirow{4}{*}{Test} & \textit{2012-02-12} & 5522 & midday & sunny \\
        & \textit{2012-02-19} & 5531 & midday & partly cloudy \\
        & \textit{2012-03-31} & 5661 & midday & cloudy \\
        & \textit{2012-05-26} & 5632 & evening & sunny  \\
      \bottomrule
    \end{tabularx}
    \begin{tablenotes}[para,flushleft]
       \footnotesize   
       The selected NCLT sequences covering diverse environmental conditions, including variations in daytime, sky, foliage, and snow. The sequences are split following LiSA~\cite{Yang_2024_CVPR}, with the upper four used for training and the lower four for validation.
     \end{tablenotes}
   \end{threeparttable}
\end{table}

\subsection{Metrics}
\label{sec:exp-metrics}
Compared to the task of VPR, which is typically evaluated only in terms of recall, we compare our method against a number of baselines both in terms of precision and recall. In more detail, we compute the Maximum Recall (MR), which constitutes the highest possible recall under a precision of 100\%, i.e., without incurring any false positive predictions. In addition, we also report the Average Precision (AP) to quantify the integral under the precision-recall curve, as done in previous loop closure detection methods~\cite{yu2024gv, merrill2018lightweight}.
In addition, we measure the predicted pose accuracy in terms of Absolute Translation Error (ATE) and Relative Pose Error (RPE). Since we do not regress the scale of the explored scene, the ATE measurements are reported up-to-scale, i.e., we normalize translation vectors to unit length and compare their L2 distance. The RPE is computed as the relative geodesic distance on $SO(3)$:
\begin{equation}
    \theta = \operatorname{cos}^{-1}\left(\frac{tr(\mathbf{R}_{gt}^T \hat{\mathbf{R}}) - 1}{2} \right),
\end{equation}
where $\mathbf{R}_{gt}$ represents the relative ground truth rotation and $\hat{\mathbf{R}}$ denotes the predicted relative camera rotation.

\subsection{Implementation Details}
\noindent{\textit{Keypoint Extraction and VLAD Clustering}:
Since our proposed LoopGNN architecture is agnostic to the choice of keypoint extractor, we evaluate its performance gain across four different traditional and learning-based extractors, namely SIFT~\cite{lowe2004sift}, SuperPoint~\cite{detone2018superpoint}, ALIKED~\cite{Zhao2023ALIKED}, and XFeat~\cite{potje2024xfeat}.
Across all keypoint types, we fit VLAD vocabularies using 64 cluster prototypes and up to 2048 keypoints per image. We use cosine-similarities for clustering in the case of XFeat, ALIKED, and SuperPoint given their deep feature embedding characteristic and Euclidean distances for SIFT. For all deep keypoint extractors, we find retrieval rates of at least 94.5\% within the top 1\% of keyframes. Nonetheless, we note that this does not imply that the true entirety of actual loops is retrieved as just 1\% of keyframes is evaluated.
}

\setlength{\tabcolsep}{3pt}
\renewcommand{\arraystretch}{1}
\begin{figure}
	\centering
    \footnotesize
    \setlength{\tabcolsep}{0.05cm}
    {\renewcommand{\arraystretch}{1}
    \resizebox{\linewidth}{!}{%
    \begin{tabular}{ccc}
  		\includegraphics[width=0.5\textwidth,clip,angle=0]{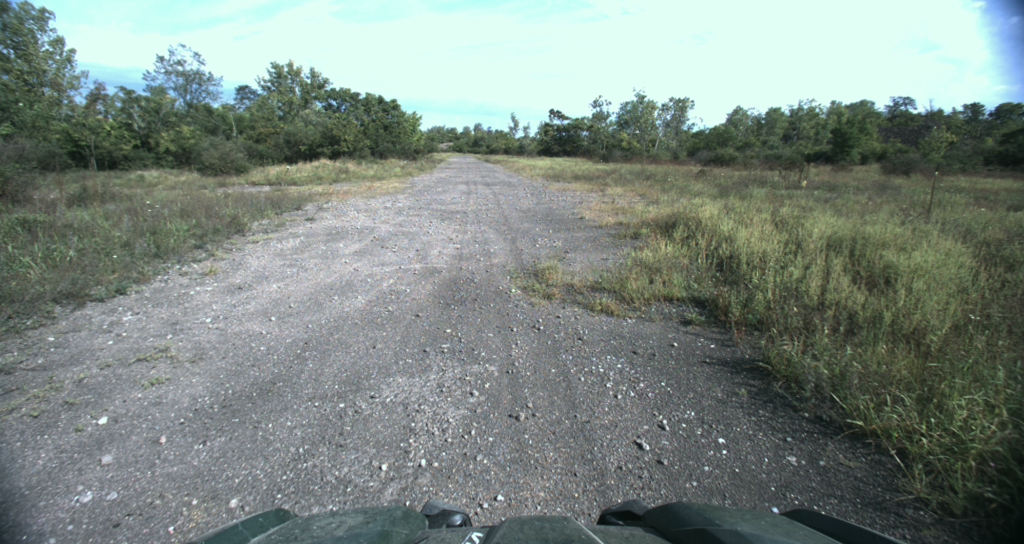} & 
        \includegraphics[width=0.5\textwidth,clip,angle=0]{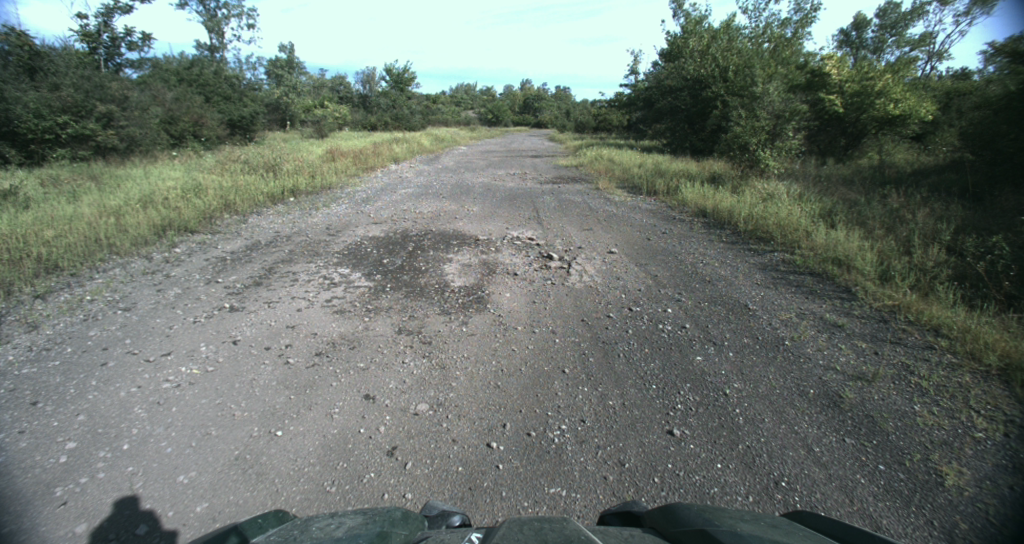} \\
        \includegraphics[width=0.5\textwidth,clip,angle=0]{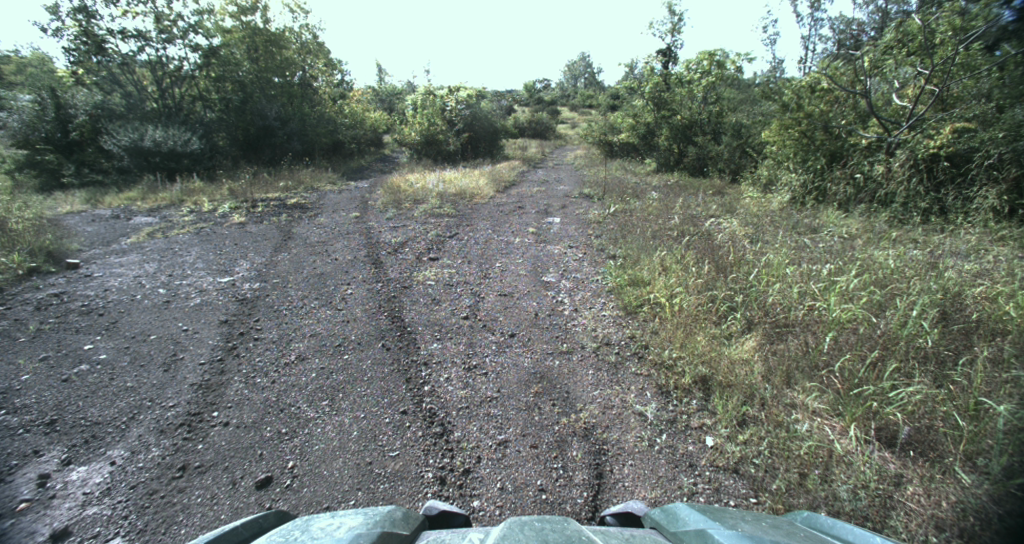} &  \includegraphics[width=0.5\textwidth,clip,angle=0]{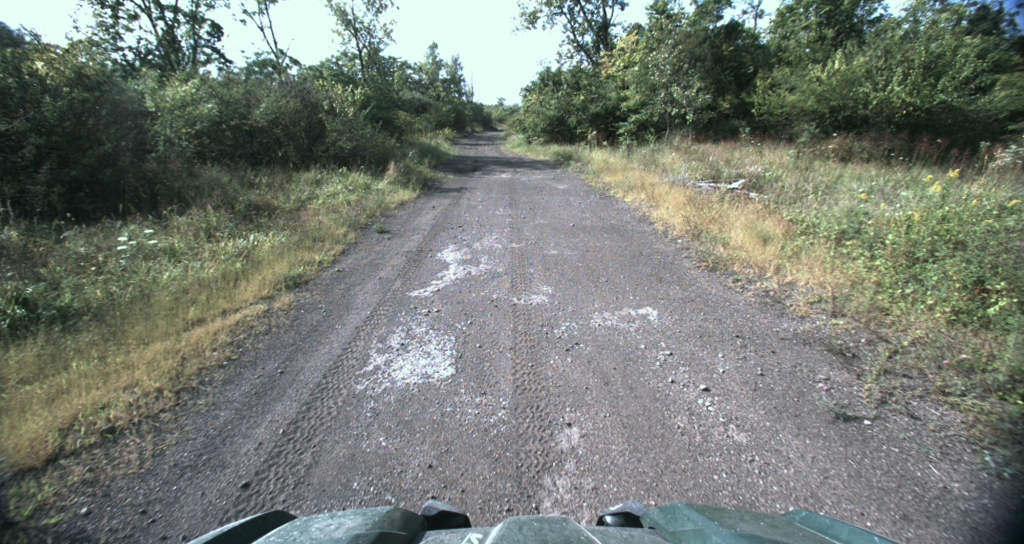} \\ \includegraphics[width=0.5\textwidth,clip,angle=0]{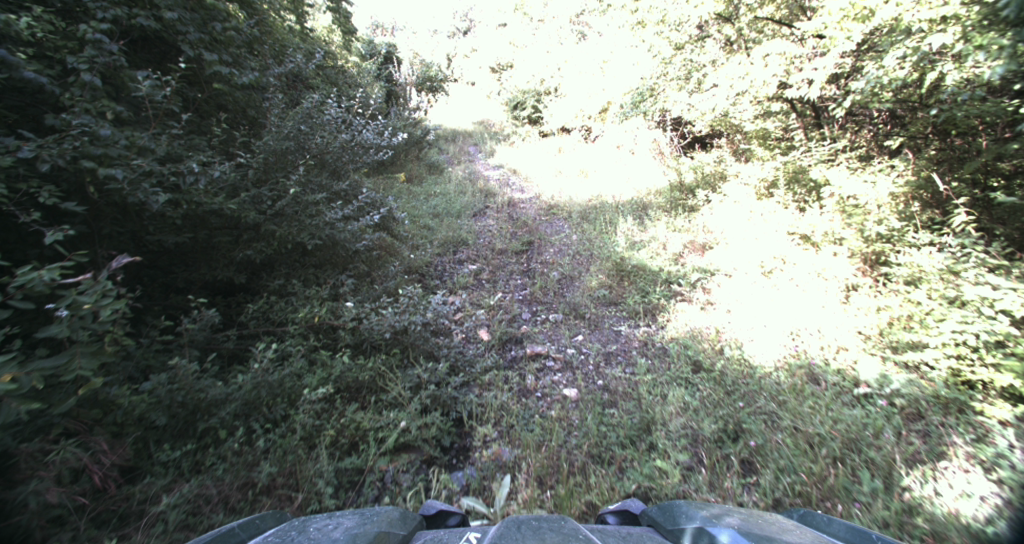} &
        \includegraphics[width=0.5\textwidth,clip,angle=0]{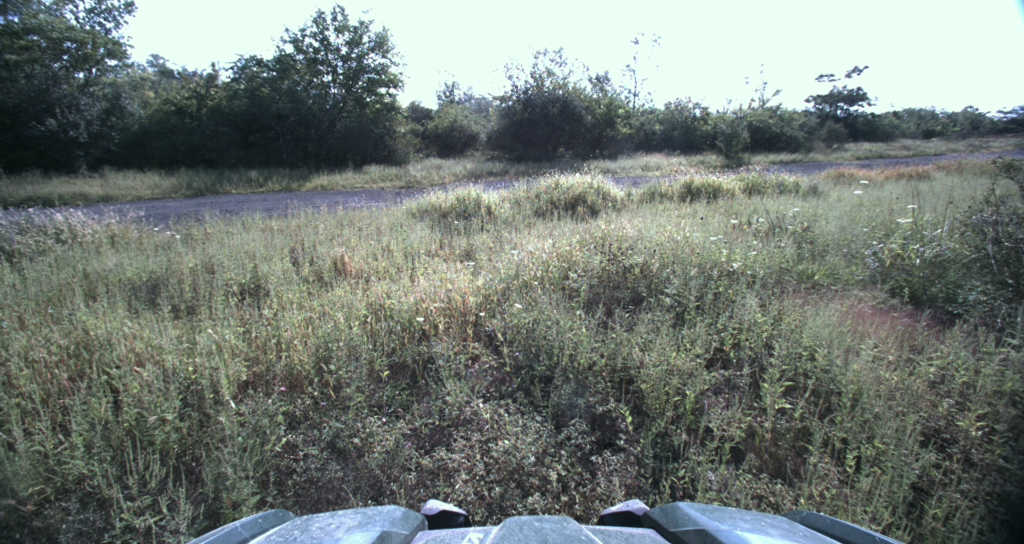} \\
    \end{tabular}}
    }
    \vspace{-0.3cm}
	\caption{A selection of various keyframes of the TD2 sequence \texttt{figure\_8\_2023-09-13-17-24-26} underlining the degree of perceptual aliases contained.}
  	\label{fig:td2-test-set}
\end{figure}
\setlength{\tabcolsep}{6pt}
\renewcommand{\arraystretch}{1}

\noindent{\textit{Network Architecture}: We train the proposed network for 10 epochs using a batch size of $2$, i.e., two retrieved neighborhoods, a learning rate of $1e-4$, and $6$ message passing steps. During training, we apply a dropout probability of $0.2$ and thus sample various node neighborhoods stochastically to increase robustness. To prevent overfitting on the training dataset, we employ early stopping based on the per-batch AP and MR.
During training, we consider both intra- and inter-sequence loops to increase robustness to lighting and viewpoint variations. Given the high amount of potential keyframe pairs, we evaluate only intra-sequence loops on the test set for both \textit{NCLT} and \textit{TartanDrive 2.0}.
}

\subsection{Baselines}
\label{exp:baselines}
We compare our proposed method LoopGNN with a set of deep loop closure detection baselines, namely Calc~\cite{merrill2018lightweight}, LoopCNN~\cite{zhang2017loop}, and our own constructed baseline Pair-NetVLAD. In addition, we examine LoopGNN in comparison to two prominent visual place recognition methods, VLAD~\cite{arandjelovic2013all} and DBoW2~\cite{galvez2012bags}. Finally, we assess the performance when employing robust keypoint matching and RANSAC for geometric verification. 

\noindent{\textit{Calc}}: 
Calc~\cite{merrill2018lightweight} uses an unsupervised deep learning framework for loop closure detection as detailed in \cref{sec:related-work}. We fine-tune the base Calc model on the TartanDrive 2.0 and NCLT training splits, respectively. \\
\noindent{\textit{LoopCNN}}:
LoopCNN~\cite{zhang2017loop} utilizes a pre-trained convolutional neural network model that is not fine-tuned on target datasets, which we make similar use of. \\
\noindent{\textit{Pair-NetVLAD}}: We present a naive baseline method called Pair-NetVLAD by eliminating the message passing stage of our proposed LoopGNN network, see \cref{sec:message-passing}. Under a GNN depth of zero, the node features do not benefit from feature propagation within neighborhoods, thus disabling the proposed graph consensus mechanism. Instead, the Pair-NetVLAD baseline produces keyframe embeddings that are stacked for each pair of keyframes and regressed using an MLP to yield a binary score per edge. \\
\noindent{\textit{XFeat + RANSAC}}: In addition to the deep loop closure detection baselines, we provide experimental results using a classical keypoint matcher relying on XFeat~\cite{potje2024xfeat} and RANSAC under an epipolar constraint for robust geometric verification, as utilized in previous studies~\cite{yu2024gv}. Since we employ XFeat's deep keypoint descriptors in LoopGNN, we believe that this comparison is representative.\\
\noindent{\textit{VLAD \& DBoW2}}: We also assess the performance of two prominent VPR methods, VLAD~\cite{arandjelovic2013all} and DBoW2~\cite{galvez2012bags}.

\subsection{Visual Loop Closure Detection}
In the following, we compare the performance of LoopGNN against the set of baselines described in \cref{exp:baselines} on the TD2 and NCLT datasets (\cref{sec:exp-datasets}). Across all experiments, we first retrieve the top-1\% neighborhoods by comparing global VLAD descriptors as stated in \cref{sec:method-global-feat-agg} and report the predictive performance only among the retrieved neighboring frames, not against all possible loops. This is due to the sheer number of possible combinations to be evaluated reaching $10^8$ keyframe pairs for the longest TD2 test sequence.

    \begin{table}
        \footnotesize
        \setlength{\tabcolsep}{7.5pt}
        \centering
        \caption{Loop Closure Detection Performance on TartanDrive 2.0}
        \label{tab:td2-evaluation}
        \begin{threeparttable}
        \begin{tabularx}{\linewidth}{ll|cccc}
          \toprule
            & \multirow{2}{*}{Model} & AP &  MR & RPE & ATE \\
            & & [\%] & [\%]  & [deg] & [1]\\
             \midrule
            \multirow{2}{*}{\rotatebox{90}{VPR}} & DBoW2~\cite{galvez2012bags} & 40.84  & 00.34 & - & - \\
             & VLAD~\cite{arandjelovic2013all} & 70.85 & 00.00 & - & - \\
             \greyrule
             \multirow{4}{*}{\rotatebox{90}{LCD}} & Calc~\cite{merrill2018lightweight} & 57.56 & 02.79 & - & - \\
            & Pair-NetVLAD & 43.38 & 06.39 & - & - \\
            & LoopCNN~\cite{zhang2017loop} & 66.26 & 01.55  & - & - \\
            & \ours{} (ours) & \textbf{84.63} & \textbf{45.52} & - & - \\
            \greyrule
            \multirow{2}{*}{\rotatebox{90}{GV}} & XFeat + RANSAC & 85.94 &  28.91 & \textbf{9.566} & \textbf{2.459} \\
            & \ours{} + RANSAC & \textbf{94.39} & \textbf{79.10} & 10.42 & 2.461 \\ 
          \bottomrule
        \end{tabularx}
        \begin{tablenotes}[para,flushleft]
           \footnotesize   
           The average precision (AP) and maximum recall (MR) are reported as average of the sequences. We compute the average translation error (ATE) as well as rotation error (RPE) under all retrieved pairs constituting a ground truth loop closure. VPR, LCD, and GV denote visual place recognition methods, deep loop closure detection methods, and geometric verification-based methods.
         \end{tablenotes}
       \end{threeparttable}
    \end{table}

In general, we regard TD2 to be the vastly more challenging dataset as it exhibits less structural features, especially on the test set, as shown in \cref{fig:td2-test-set}. As detailed in \cref{tab:td2-evaluation}, we observe that the two VPR methods, DBoW2 using ORB keypoints~\cite{rublee2011orb} and VLAD utilizing XFeat features~\cite{potje2024xfeat} generally yield relatively high average precision given that they are optimized towards high recalls. However, we note that both methods produce subpar maximum recalls at precision equal to 100\%. As expected, this indicates that VPR methods are not robust to perceptual aliases and require downstream geometric verification to produce accurate loop closures.

Next, we compare various deep loop closure detection methods in \cref{tab:td2-evaluation}. While our baseline PairNet-VLAD underperforms compared to both Calc and LoopCNN in terms of average precision, it yields a slightly higher maximum recall. We attribute this to the direct binary supervision the network benefits from, whereas the two other methods directly maximize feature similarity given an external learning signal. In contrast, LoopGNN operates on VLAD-retrieved neighborhoods of frames, which drastically improves both the average precision and recall as detailed in \cref{tab:td2-evaluation}, thus being an order of magnitude more robust towards false positive detection of perceptual aliases.

Lastly, we compare LoopGNN against a keypoint matcher utilizing XFeat's keypoints followed by a geometric verification scheme. In addition, we also report the accuracy of the corresponding relative camera transformations in \cref{tab:td2-evaluation} using the RPE and ATE metrics, see ~\cref{sec:exp-metrics}. We observe that XFeat + RANSAC outperforms the standard LoopGNN in terms of AP. Nonetheless, this comes at the cost of a drastic decrease in maximum recall at 100\% precision. This indicates that the XFeat + RANSAC pipeline is not robust to large amounts of perceptual aliases as depicted in \cref{fig:td2-test-set}. Nevertheless, it's MR of 28.91\% is substantially higher than the remaining deep loop closure detection and VPR baselines.
In contrast, when combining LoopGNN with RANSAC, by geometrically verifying the top-0.5\% of scores predicted by LoopGNN using RANSAC, we observe an AP increase of around 10\% while reaching maximum recalls of nearly 80\%.
This ultimately demonstrates the robustness of LoopGNN in identifying perceptual aliases through deep graph consensus while introducing a minimal computational overhead.
In addition to the overall quantitative results on TD2, we visualize the predictions of XFeat + RANSAC and LoopGNN in \cref{fig:qual-comparison}. Although LoopGNN yields robust estimates, we observe that the top-1\% VLAD neighborhoods retrieve only a fraction of the true loop closures under a high degree of perceptual aliases, thus representing a significant bottleneck in loop closure detection.

\begin{figure}
    \centering
    \footnotesize
    \includegraphics[width=0.24\textwidth,trim={1.6cm 1cm 0 0},clip]{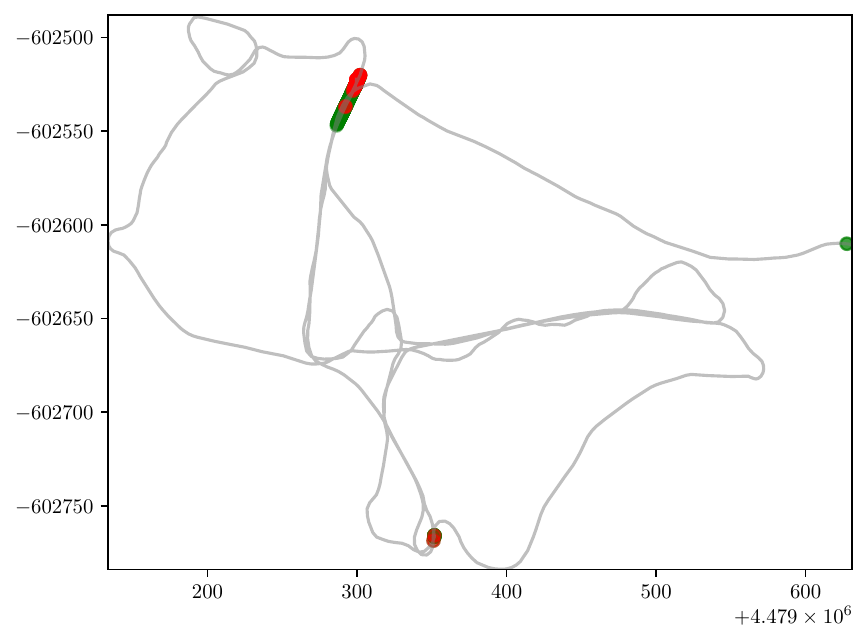}
    \includegraphics[width=0.24\textwidth,trim={1.6cm 1cm 0 0},clip]{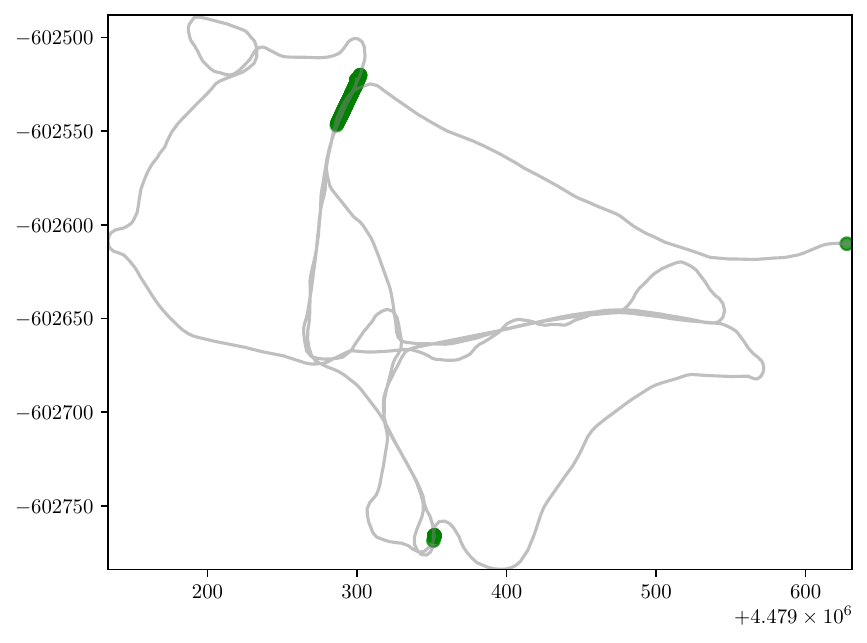}
    \caption{Qualitative results comparing the loop closure predictions of XFeat~\cite{potje2024xfeat} + RANSAC (left) against LoopGNN (right) under similar VLAD neighborhood retrieval of 1\% at maximum recall at 100\% precision. Red dots represent false negative loop closures, while green dots represent true positive loop closures. The underlying scene is \texttt{figure\_8\_2023-09-13-17-24-26} of the TartanDrive 2.0 dataset.
    }
    \label{fig:qual-comparison}
\end{figure}

\begin{table}
    \footnotesize
    \setlength{\tabcolsep}{25pt}
    \centering
    \caption{Loop Closure Detection Performance on the NCLT Dataset}
    \label{tab:nclt-evaluation}
    \begin{threeparttable}
    \begin{tabularx}{1.0\linewidth}{l|cccc}
      \toprule
        \multirow{2}{*}{Model} & AP & MR \\
         & [\%] & [\%] \\
         \midrule
        Calc~\cite{merrill2018lightweight} & 38.71 & 00.32\\
        Pairwise-VLAD & 36.90 & 04.29 \\ 
        LoopCNN~\cite{zhang2017loop} & 50.05 & 00.92 \\
        \ours{} (ours) & \textbf{71.69} & \textbf{20.82} \\ 
      \bottomrule
    \end{tabularx}
    \begin{tablenotes}[para,flushleft]
       \footnotesize
       2\% retrieval ratio
       The average precision (AP) and maximum recall (MR) are evaluated per-sequence.
       We compute the average translation error (ATE) as well as rotation error (RPE) using given ground-truth loop closures or all true positive predicted loop closures. 
     \end{tablenotes}
   \end{threeparttable}
   \vspace{-0.3cm}
\end{table}

Additionally, we train LoopGNN on the NCLT dataset and compare it against Calc~\cite{merrill2018lightweight}, LoopCNN~\cite{zhang2017loop}, and our Pair-NetVLAD baseline. Different to the TD2 dataset, we opt for a larger neighborhood size of 2\% given the shorter sequence length of the reprocessed NCLT dataset. As detailed in \cref{tab:nclt-evaluation}, LoopGNN outperforms both Calc and LoopCNN in terms of AP and MR, thus showing a similar trend as on the TD2 data. Compared to its Pair-NetVLAD counterpart, LoopGNN is more robust.

\begin{table}
    \footnotesize
    \setlength{\tabcolsep}{13pt}
    \centering
    \caption{Ablation Study over Various Keypoint Extractors}
    \label{tab:td2-keypoint-ablation}
    \begin{threeparttable}
    \begin{tabularx}{\linewidth}{ll|cc}
      \toprule
        Feature & Model & AP & MR \\
        Extractor & Variant & [\%] & [\%] \\
         \midrule
            \multirow{2}{*}{SIFT~\cite{lowe2004sift}} & Pair-NetVLAD & 00.00 & 00.00 \\
            & LoopGNN & 00.00 & 00.00 \\
            \greyrule
            \multirow{2}{*}{SuperPoint~\cite{detone2018superpoint}} & Pair-NetVLAD & 57.18 & 07.14 \\ 
            & LoopGNN & \textbf{84.49} & \textbf{21.00} \\
            \greyrule
            \multirow{2}{*}{ALIKED~\cite{edstedt2024roma}} & Pair-NetVLAD & 57.67 & 12.85\\ 
            & LoopGNN & \textbf{78.83} & \textbf{30.74} \\ 
            \greyrule
            \multirow{2}{*}{XFeat~\cite{potje2024xfeat}} & Pair-NetVLAD & 43.38 & 06.39 \\
            & LoopGNN & \textbf{84.63} & \textbf{45.52} \\
      \bottomrule
    \end{tabularx}
    \begin{tablenotes}[para,flushleft]
       \footnotesize      
       Across all experiments, we retrieve the 1\% of closest keyframes given a query keyframe using fitted VLAD clusters under the respective keypoint type mentioned. We demonstrate that our method shows significant performance gains in both AP and MR when employing any kind of deep keypoint feature extractor (SuperPoint, ALIKED, XFeat). On the contrary, we observe that within our framework traditional keypoint methods such as SIFT are not producing valuable estimates when used in VLAD clustering as well as loop closure detection.
     \end{tablenotes}
   \end{threeparttable}
\end{table}

\begin{table}
    \footnotesize
    \setlength{\tabcolsep}{20pt}
    \centering
    \caption{Ablation Study of Neighborhood Sizes}
    \label{tab:td2-neighbor-ablation}
    \begin{threeparttable}
    \begin{tabularx}{1.0\linewidth}{r|cc}
      \toprule
        Neighborhood Size & AP [\%]  & MR [\%] \\
         \midrule
            0.5\%  & 80.03 & \underline{17.34} \\
            1.0\% & \textbf{84.63}  & \textbf{45.52} \\
            1.5\% & \underline{80.39} & 16.44 \\
      \bottomrule
    \end{tabularx}
    \begin{tablenotes}[para,flushleft]
       \footnotesize      
       We report the average precision and maximum recall at precision equal to one (MR) across various retrieved neighborhood sizes fed to the LoopGNN network.
     \end{tablenotes}
   \end{threeparttable}
\end{table}

\subsection{Ablation Study}
In the following, we ablate on two key design parameters of our proposed framework. First, we demonstrate that LoopGNN's capabilities are orthogonal to the underlying deep keypoint descriptor as detailed in \cref{tab:td2-keypoint-ablation}. Across all three deep keypoint methods (SuperPoint\cite{detone2018superpoint}, ALIKED~\cite{Zhao2023ALIKED}, XFeat~\cite{potje2024xfeat}), we observe a similar relative performance improvement over the Pair-NetVLAD baseline. However, under traditional keypoint descriptors such as SIFT~\cite{lowe2004sift} our method does not produce viable outputs. We attribute this to significantly lower recalls during the VLAD retrieval stage and lower fidelity of SIFT features.

Furthermore, we demonstrate the effect of different neighborhood sizes on the performance of LoopGNN in \cref{tab:td2-neighbor-ablation}. In general, we notice small decreases in terms of AP while the robustness to false positives measured as MR drops sharply. Thus, adequately-sized VLAD neighborhoods have a substantial influence on prediction robustness.

\subsection{Computational Considerations}
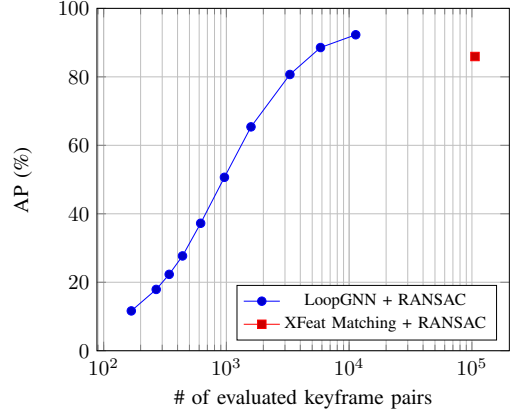
\begin{figure}
    \centering
    \begin{tikzpicture}[scale=0.8]
    \begin{axis}[
        xlabel={\# of evaluated keyframe pairs},
        ylabel={AP (\%)},
        xmode=log,
        log basis x=10,
        ymin=0, ymax=100,
        grid=both,
        legend pos=south east,
        legend style={font=\footnotesize}
    ]
    \addplot coordinates {
         (168, 11.63) (268, 17.91) (342, 22.31) (439, 27.69) (617, 37.23) (964, 50.66) (1586, 65.39) (3291, 80.69) (5842, 88.56) (11309, 92.29)
    };
    \addplot coordinates {
        (105895, 85.94)
    };
    \addlegendentry{LoopGNN + RANSAC}
    \addlegendentry{XFeat Matching + RANSAC}
    
    \end{axis}
    \end{tikzpicture}
    \vspace{-0.3cm}
    \caption{Efficiency comparison of LoopGNN employing XFeat keypoint descriptors against classical XFeat descriptor matching on \textit{TartanDrive 2.0}. The matches of both methods are geometrically verified under the epipolar model to increase robustness. Both methods draw keyframe pairs from the top-1\% VLAD neighborhoods on TD2 scene \texttt{figure\_8\_2023-09-13-17-24-26}.
    }
    \vspace{-0.3cm}
    \label{fig:efficiency-plot}
\end{figure}
We demonstrate the computational benefits of running LoopGNN in comparison to utilizing a classical deep image matching + RANSAC pipeline as evaluated in \cref{tab:td2-evaluation}. In \cref{fig:efficiency-plot}, we present the computational efficiency of LoopGNN over a classical matching pipeline. We observe a consistently rising AP when gradually increasing the number of considered top-scoring candidate edges from $10^2$ to $10^4$ for a scene of 4400 keyframes. Thus, LoopGNN allows for a considerable reduction in the number of keyframe pairs requiring geometric verification. While it is generally not feasible to verify large numbers of candidates in an online scenario, LoopGNN provides precise scoring to ultimately decrease the number of candidates that need to undergo verification. In addition, GPU-based inference while retrieving top-1\% neighborhoods allows for inference speeds of up to \si{20}{Hz} on a NVIDIA A6000 GPU. Given these findings, we believe that integrating LoopGNN into an online visual SLAM method allows for a considerable reduction in map drift due to higher precision and recall of detected loop closures.

\section{Conclusion}
In this work, we introduced LoopGNN, a novel deep graph consensus approach to estimate robust loop closures. We found that conducting deep information exchange among retrieved initial candidates allows for a significant reduction in the actual number of keyframes that need to undergo geometric verification as demonstrated on two distinct datasets. In future work, we intend to evaluate the feasibility of larger neighborhoods and aim to integrate a differentiable RANSAC module into the LoopGNN architecture to perform batched geometric verification within a single forward pass. To foster future research in this domain, we make code and reprocessed keyframe data publicly available.


\footnotesize
\bibliographystyle{IEEEtran}
\bibliography{bibliography.bib}



\end{document}